\documentclass{bmvc2k}

\usepackage{amssymb}
\usepackage{booktabs}
\usepackage[normalem]{ulem}
\useunder{\uline}{\ul}{}

\title{Rethinking Microscopy Generation: Co-Designed Diffusion for Biologically Interpretable Single-Cell Synthesis}

\addauthor{Shumpei Takezaki$^{*}$}{shumpei.takezaki@human.ait.kyushu-u.ac.jp}{1}
\addauthor{Duway Nicolas Lesmes-Leon$^{*}$}{duway_nicolas.lesmes_leon@dfki.de}{2,3}
\addauthor{Ryoma Bise}{bise@ait.kyushu-u.ac.jp}{1}
\addauthor{Gillian Lovell}{gillian.lovell@sartorius.com}{4}
\addauthor{Bianca Migliori}{bianca.migliori@sartorius.com}{4}
\addauthor{Andreas Dengel}{andreas.dengel@dfki.de}{2,3}
\addauthor{Sheraz Ahmed}{sheraz.ahmed@dfki.de}{3}

\addinstitution{
 Kyushu University, Fukuoka, Japan
}
\addinstitution{
University of Kaiserslautern-Landau\\  (RPTU), Kaiserslautern, Germany
}
\addinstitution{
German Research Center for Artificial Intelligence (DFKI),\\ Kaiserslautern, Germany
}
\addinstitution{
Sartorius, BioAnalytics, Royston, United Kingdom
}

\runninghead{TAKEZAKI ET AL.}{Rethinking Microscopy Generation}

\begin{document}

\maketitle

\begingroup
\renewcommand{\thefootnote}{\fnsymbol{footnote}}
\footnotetext[1]{Equal contribution.}
\endgroup

\begin{abstract}
Generative models for microscopy can already produce realistic whole-image fields of view, but offer no principled way to control how each individual cell looks. We address this gap by recasting microscopy generation at the single-cell level and designing the input controls and the model's internal representation jointly, so that both are organized along the same biologically meaningful directions. On the input side, we aggregate a long list of standard cell measurements into a small set of semantic axes (size, elongation, brightness, texture, etc.) that are each named, measurable, and traversable. On the model side, we introduce BioVAE, an encoder-decoder whose latent space is explicitly aligned with the same set of measurements. A key finding is that the benefit of biologically structuring the latent emerges only when conditioning is also moved to the semantic axes, showing that the two ends must be designed jointly. Experiments on LIVECell show that semantic-axis conditioning substantially improves generative fidelity over high-dimensional conditioning, a blind study with biologists confirms the realism of generated cells, and per-axis traversals produce stable and interpretable phenotypic changes.
\end{abstract}

\begin{figure}[t]
    \centering
    \includegraphics[width=\linewidth]{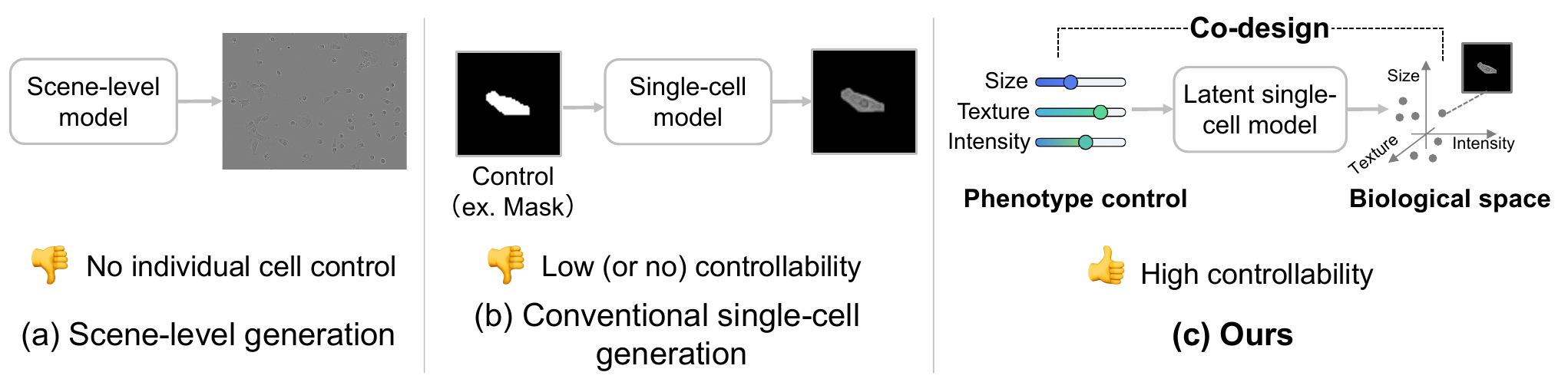}
    \caption{Rethinking phenotype control in microscopy generation. (a) Scene-level methods generate the whole field of view and cannot directly control individual cells. (b) Existing single-cell methods do not jointly design the conditioning signal and the latent representation, so phenotype control is unreliable or out of reach. (c) Our framework co-designs both around biologically meaningful semantic axes, enabling stable and interpretable per-cell control.}
    \label{fig:concept}
\end{figure}

\section{Introduction}
Microscopy images are pictures of a set of individual cells captured under a microscope, and the way each cell looks; its size, shape, brightness, and internal texture patterns, is defined as the phenotype. Phenotypic differences across cells are the basis of almost every quantitative question in cell biology, from how cells respond to a drug to how diseased cells differ from healthy ones~\cite{data_nature,cellHeterogeneity}. Recent generative models can already generate realistic microscopy images~\cite{class_condition,CellGAN,I2I_app,eschweiler2024celldiffusion,eschweiler2021gan,Yu2023diffusion}, but they produce the entire field-of-view in one shot rather than building it cell by cell (Fig.~\ref{fig:concept}~(a)). As a result, current models do not provide direct control over interpretable phenotype attributes such as cell size or brightness. While cellular phenotype is fundamentally defined at the level of individual cells, current generative models typically operate at the level of entire microscopy scenes.

These limitations motivate studying generation at the level of individual cells. Fine-grained control over cellular phenotype naturally suggests formulating generation at the level of individual cells. The remaining challenge is how to specify desired cell phenotypes in a form that is both biologically meaningful and amenable to reliable generation.

Several existing methods can already generate realistic single-cell images~\cite{VAE+,CP2Img,cascadad_diff,Goldsborough227645}, but fine-grained control over phenotype remains challenging. Some methods condition generation on binary shape masks and synthesize the corresponding cell appearance (Fig.~\ref{fig:concept}~(b)); the user can specify the cell outline but has no control over its brightness or texture. Other methods take a long list of raw features measuring cell properties as input, but many of them describe overlapping aspects of the same cell, and many are not faithfully represented in the generated image. In both cases, precise control over individual phenotype attributes remains limited, making it difficult to manipulate generated cells along isolated biologically meaningful directions such as size or brightness.

We address this challenge through a joint design of the conditioning space and latent representation, both organized around the same biologically meaningful phenotype directions (Fig.~\ref{fig:concept}~(c)). On the input side, instead of asking the user for a shape mask or a long list of raw features, we provide a small set of intuitive controls, one for each interpretable aspect of how a cell looks, such as size, elongation, brightness, and texture. We call these the semantic axes. On the output side, the model internally compresses each cell image into a compact code before generating it, and we train this compression so that the code itself is organized along the same set of biologically meaningful directions. We call this component BioVAE. Aligning the conditioning variables and latent representation around the same phenotype directions promotes stable and predictable control over generated cells. To our knowledge, this joint design has not been studied for single-cell image generation.

Our contribution is this joint design rather than single-cell synthesis itself, which has been demonstrated in prior work~\cite{CP2Img, cascadad_diff}. We also see single-cell and whole-image generation as complementary: whole-image models capture how cells are arranged together in a microscope picture, while controllable single-cell generation provides a fine-grained handle on how each individual cell looks. Together they form a more complete toolkit for studying cellular phenotype with generative models.

\begin{figure}[t]
    \centering
    \includegraphics[width=\linewidth]{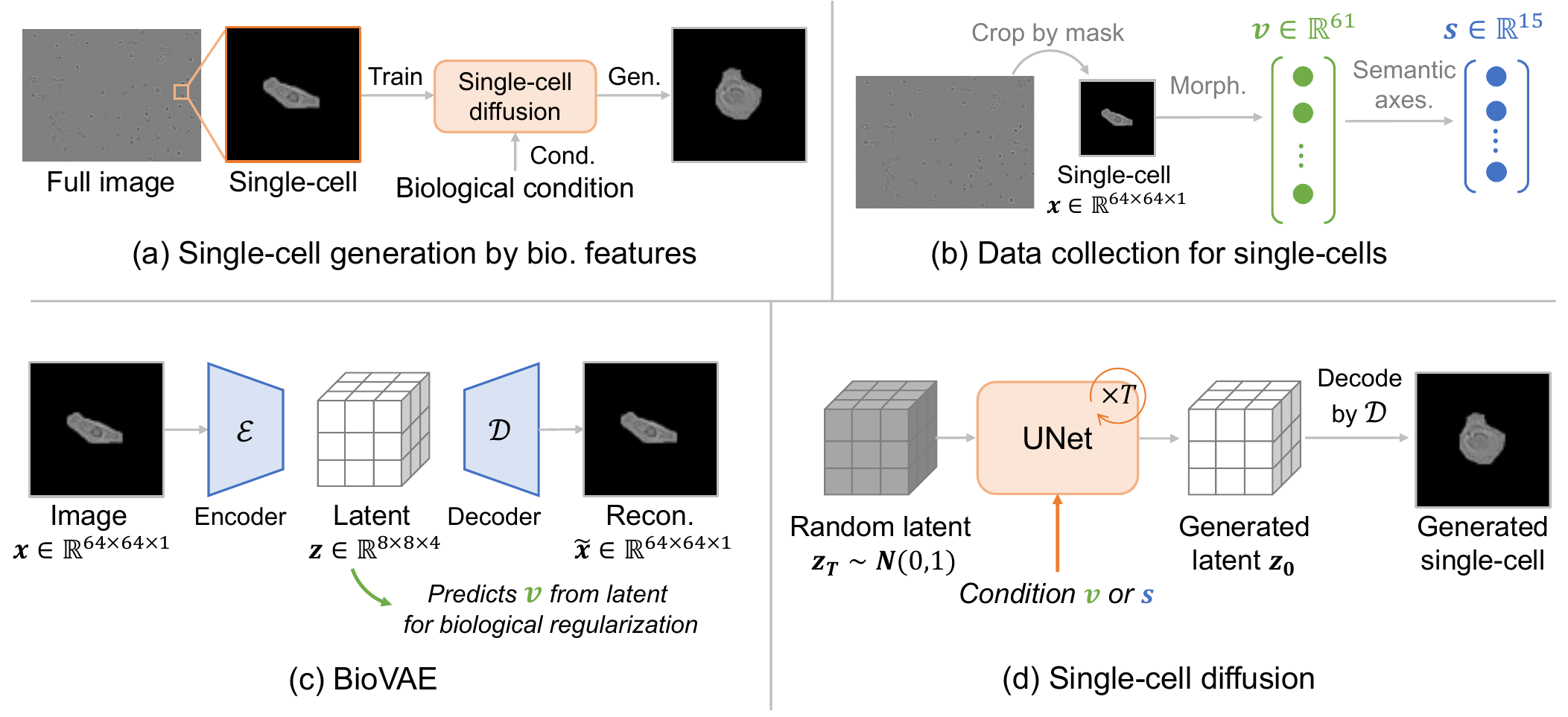}
    \caption{Overview of the controllable single-cell latent diffusion framework. (a) A concept of single-cell generation. (b) Single-cell extraction from the LIVECell~\cite{livecell} dataset and calculation of morphodynamic descriptors ($v$) and semantic axes ($s$). (c) The Biologically Regularized VAE (BioVAE) architecture, which predicts $v$ from the latent space to enforce biological structure. (d) The conditional single-cell latent diffusion model guided by morphodynamic descriptors $v$ or semantic axes $s$ for controllable synthesis.}
    \label{fig:method}
\end{figure}

In this work, we propose a controllable single-cell generation framework whose conditioning variables and latent representation are jointly organized around biologically meaningful phenotype directions (Fig.~\ref{fig:method}~(a)). We first extract standard measurements of cellular appearance~\cite{Biofeats} and aggregate them into a compact set of semantic axes that serve as interpretable conditioning variables (Fig.~\ref{fig:method}~(b)). To align generation with these controls, we introduce BioVAE, an encoder-decoder whose latent representation is explicitly structured around the same phenotype measurements (Fig.~\ref{fig:method}~(c)). A latent diffusion model~\cite{ldm} is then trained within this representation space and conditioned on the semantic axes, enabling controllable generation of individual cells (Fig.~\ref{fig:method}~(d)).

Experiments on the LIVECell~\cite{livecell} dataset demonstrate that semantic-axis conditioning provides more effective control than conditioning on the raw set of measurements while maintaining high generation quality. A blind study with biologists further supports the realism of the generated cells, and traversal experiments show that individual semantic axes correspond to stable and interpretable phenotypic variations.

Our contributions are summarized as follows:
\begin{itemize}
\item \textbf{Problem framing.} We articulate controllable single-cell generation as a complementary direction to whole-image microscopy generation, and identify the joint design of the user-facing controls and the model's internal representation as the central methodological question at this scale.
\item \textbf{BioVAE.} We propose an encoder-decoder whose compressed representation is explicitly aligned with biological measurements of how cells look, via an auxiliary regression objective and an anti-collapse entropy term.
\item \textbf{Semantic-axis conditioning.} We introduce a compact set of semantic axes, where each axis is named (corresponds to a recognizable phenotype), measurable (can be re-evaluated from the generated image), and traversable (can be varied independently), and show that this works better than conditioning on the raw list of measurements.
\item \textbf{Empirical validation.} On LIVECell, we report generation quality, a blind study with biologists, and per-axis controllability via traversal experiments.
\end{itemize}

\section{Related work}

\subsection{Whole-Image Microscopy Generation}
Most generative work in microscopy produces the entire picture in one shot. Early GAN-based methods synthesize whole fluorescence images, typically conditioned on coarse labels such as cell type or treatment, or on annotation masks of multi-cellular structures~\cite{eschweiler2021gan, CellGAN}. More recent diffusion-based methods generate or restore whole microscopy fields~\cite{eschweiler2024celldiffusion, Yu2023diffusion, I2I_app, class_condition}, sometimes combined with mask-to-image translation or single-cell composition. These methods capture global properties of a microscope picture well, such as how densely cells are packed and how they are arranged in space, but they cannot specify how any individual cell should look, because the conditioning signal describes the whole field of view rather than each cell.

\subsection{Single-Cell Modeling and Synthesis}
A complementary line of work models individual cells, but each leaves a controllability gap that our framework targets. Early GAN-based methods such as CytoGAN~\cite{Goldsborough227645} generate single-cell images without any conditioning, and so offer no control over how the generated cell looks. CP2Image~\cite{CP2Img} takes a different approach: the user provides a long vector of standard measurements of cell appearance (the CellProfiler~\cite{cellprofiler} representation, on the order of several hundred numbers), and the model generates a matching cell image. In principle this allows the user to specify a target appearance, but in practice the measurements are highly correlated, and \cite{CP2Img} reports that fewer than a third of them are faithfully reflected in the generated image. As a result, the user cannot reliably make the generated cell, for example, larger or brighter on demand. Cascaded diffusion for cell synthesis~\cite{cascadad_diff} takes yet another route: it samples a binary cell-shape mask and then fills it in with a fine-tuned Stable Diffusion model. The mask sampler is unconditional, so the user cannot specify the shape, and once a mask is given everything else (brightness, texture, internal structure) is left to the diffusion model; the pipeline is designed for generating training data for cell segmentation rather than for controlling phenotype. Unsupervised single-cell representation learning~\cite{VAE+} learns compact representations of cell images, but does not generate. In all of these methods, the input the user can manipulate and the model's internal representation are designed independently, with no shared set of interpretable directions tied to how cells look. We design both together so that a small set of intuitive controls (size, elongation, brightness, texture) drives both ends, making per-cell phenotype controllable in a stable and interpretable way.

\subsection{Controllable and Conditional Diffusion}
Conditional diffusion models steer generation through mechanisms such as cross-attention conditioning~\cite{ldm}, feature-wise modulation (FiLM)~\cite{film}, and classifier-free guidance~\cite{cfg}. These mechanisms determine \textit{how} a conditioning signal is injected into the generator, but they do not address \textit{what} that signal should be or how it should relate to the model's internal representation. In the single-cell setting, the natural conditioning vocabulary is a long list of correlated handcrafted measurements, and we find empirically that feeding this redundant signal directly into a standard FiLM-conditioned diffusion model leads to unstable generation and unreliable per-axis control (Sec.~\ref{gen_qual}). Our semantic-axis design, together with the BioVAE latent it is aligned with, can be viewed as a domain-specific co-design of \textit{what} to condition on and \textit{what} the latent represents, which lets an otherwise standard conditional diffusion model behave stably and interpretably.

\subsection{Disentangled and Structured Latent Spaces}
A long line of work asks how a generative model can be made interpretable, in the sense that each direction in its latent space corresponds to one clean, isolated change in the output (for example, only the size of an object, or only its color). Some methods train the model so that different latent dimensions are statistically independent of one another~\cite{betavae}. Others take an already trained generator and search inside its latent space to find directions that happen to correspond to meaningful changes~\cite{ganspace}. In both cases, the criterion for a good direction is statistical independence: moving along one direction should not affect the others.

Our setting differs in two respects. First, our directions are not discovered by the model: each one is defined in advance from a standard measurement of how cells look (size, elongation, brightness, texture), so the directions are interpretable by design. Second, our directions are intentionally not statistically independent. They are derived from groups of measurements that naturally overlap in real cells; for example, cell area, perimeter, and equivalent diameter all describe cell size and tend to move together. Forcing such directions to be independent would erase patterns that actually exist in real cells. We therefore aim for each direction to produce a change that matches how real cells vary, rather than a strictly isolated one (Sec.~\ref{sec:control}). This positions our work between classical handcrafted descriptors of cell appearance and unsupervised disentanglement: we prioritize matching real cell variation over strict statistical independence.

\subsection{Biological Descriptors in Computational Cell Biology}
Researchers studying cells have a long tradition of extracting numbers that summarize how each cell looks in a microscopy image, such as its size, shape, brightness, and the pattern of its internal texture. CellProfiler~\cite{cellprofiler} is a widely used software that automatically computes hundreds of such measurements from a cell image, and the Cell Painting protocol~\cite{Biofeats} is a standardized way of staining cells with fluorescent dyes so that these measurements consistently capture the same set of cellular features across experiments. Together they are routinely used in large-scale experiments that test many drugs or genetic conditions on cells at once. Subsequent work has shown that, although these measurements are high-dimensional, they can be summarized by a small number of meaningful patterns that capture the main ways in which cells differ from one another~\cite{caicedo2017}. We adopt these measurements not as the end product of an analysis, but as supervision and conditioning signals for image generation, connecting this established way of describing cells quantitatively with modern generative models.

\section{Controllable Single-Cell Latent Diffusion Framework}

Fig.~\ref{fig:method}~(a) illustrates the core idea of our framework: we generate individual cell images and design the input controls and the internal representation together so that both are organized along the same set of biologically meaningful directions. The framework has two components. First, on the input side, we take the long list of measurements that describes how each cell looks and aggregate them into a small set of intuitive semantic axes (size, elongation, brightness, texture, etc.); these axes are then used to condition a latent diffusion model (Fig.~\ref{fig:method}~(b,d)). Second, on the model side, we train an encoder-decoder, BioVAE, whose latent representation is explicitly aligned with the same set of measurements, so that the latent is organized along the same directions the user controls (Fig.~\ref{fig:method}~(c)). Because both ends share a common biological vocabulary, moving any control produces a stable and interpretable change in the generated cell.

\subsection{Single-Cell Extraction and Feature Aggregation}
We isolate individual cells from full microscopy images using the LIVECell~\cite{livecell} dataset and its segmentation masks. For each cell, we set the background to black, discard cells touching the image edge, and obtain a $64\times64$ single-cell crop. For each crop we then compute a long vector of standard measurements that summarize how the cell looks ($v \in \mathbb{R}^{61}$), and aggregate these into a small set of intuitive semantic axes ($s \in \mathbb{R}^{15}$) that we will use to control generation.

\noindent\textbf{Descriptor extraction.} For each cell crop, we compute 61 standard measurements inside the segmentation mask. Following established practice in computational cell biology~\cite{Biofeats,cellprofiler}, these measurements are organized into four categories that describe complementary aspects of how a cell looks: \textbf{(i) Morphology (20):} area, perimeter, equivalent diameter, major/minor axis length, aspect ratio, eccentricity, solidity, circularity, convex area, extent, and 9 Zernike moments (which describe the overall shape using a standard set of basis functions). \textbf{(ii) Pixel intensity (10):} mean, std., min, max, sum, median, and 5th/25th/75th/95th percentiles of intensities inside the mask. \textbf{(iii) Texture (25):} 13 Haralick features (which summarize the spatial pattern of intensities), a 10-bin LBP histogram (which captures local texture patterns), and gradient mean/std. \textbf{(iv) Spatial distribution (6):} a 6-bin radial intensity profile, which describes how brightness changes from the cell center outwards. Intensity-related measurements describe the cell body only, since the background is masked out. Definitions and exact formulas are provided in Supplementary Sec.~A. Let $v \in \mathbb{R}^{61}$ denote the resulting measurement vector.

\noindent\textbf{From descriptors to semantic axes.} Our goal here is not simply to reduce the 61 measurements to fewer numbers. Generic dimensionality reduction techniques such as PCA or an autoencoder would also produce a compact code, but the resulting axes would not correspond to anything that a biologist can name, manipulate, or measure independently, and so they could not serve as a control interface for phenotype. We instead manually design a small set of axes that satisfy three properties needed for interpretable control. First, each axis corresponds to a \textit{named} phenotype that a biologist would recognize (e.g., cell size, elongation, brightness, granularity). Second, each axis is \textit{measurable} from a generated image using the same pipeline that defines it, so per-axis control can be verified quantitatively after generation. Third, each axis supports \textit{one-axis traversal}: the user can vary a single axis while keeping the others fixed and observe a focused change in the generated cell. These properties are what make the axes a control interface, not just a compressed representation.

Concretely, we aggregate the 61 measurements into 15 semantic axes $s \in \mathbb{R}^{15}$, each summarizing one coherent aspect of cell appearance: \textbf{(i) Morphology (6 axes):} cell size, elongation, shape compactness, boundary irregularity, global shape complexity, and symmetry. \textbf{(ii) Pixel intensity (3 axes):} overall signal level, intensity contrast, and intensity skewness. \textbf{(iii) Texture (4 axes):} heterogeneity, smoothness, granularity, and edge strength. \textbf{(iv) Spatial distribution (2 axes):} central enrichment and radial uniformity. Each axis is computed by a simple interpretable function of the corresponding measurements (e.g., means of z-scored measurements and $L_1$ norms of Zernike moments) and normalized using training-set statistics. The complete mapping appears in Supplementary Sec.~B. The 15 axes remain partially correlated because real cells naturally vary in correlated ways (Supplementary Sec.~B.5); we do not orthogonalize them, since that would discard real patterns in cell appearance. What matters is that each axis is named, measurable, and traversable, so that moving along any axis corresponds to a biologically meaningful change in the generated cell.

\subsection{BioVAE: Biologically Regularized VAE}\label{sec:bio-reg}
The second half of the co-design is the latent representation that the diffusion model operates on. We build upon a standard VAE, which consists of an encoder $\mathcal{E}$ and a decoder $\mathcal{D}$: given an input image $x$, the encoder compresses it into a latent code $z = \mathcal{E}(x)$, and the decoder reconstructs it as $\tilde{x} = \mathcal{D}(z)$. Modern VAE architectures are trained by minimizing $\mathcal{L}_\mathrm{VAE} = \lambda_\mathrm{Recon}\mathcal{L}_\mathrm{Recon} + \lambda_\mathrm{KL}\mathcal{L}_\mathrm{KL} + \lambda_\mathrm{GAN}\mathcal{L}_\mathrm{GAN}$, where $\mathcal{L}_\mathrm{Recon}$ combines the $L_1$ loss and perceptual loss, $\mathcal{L}_\mathrm{KL}$ is the Kullback-Leibler regularization between $z$ and a standard Gaussian distribution, and $\mathcal{L}_\mathrm{GAN}$ represents the adversarial objective, with $\lambda_{(\cdot)}$ denoting the respective loss weights.

A standard VAE has no reason to organize its latent space along any particular biological direction: it only knows how to reconstruct images. We therefore introduce BioVAE, which adds a simple auxiliary objective that forces the latent to encode the same measurements that define our user-facing controls. Concretely, a shallow neural network $\mathcal{F}$ takes the latent $z$ and predicts the full measurement vector $v \in \mathbb{R}^{d}$ ($d=61$), and we minimize the mean squared error $\mathcal{L}_\mathrm{Bio} = ||\mathcal{F}(z) - v||^2 / d$. We regress the full 61-dimensional $v$ rather than the 15 aggregated axes $s$ because the latent should retain fine-grained measurement structure to serve as a general-purpose backbone, whereas $s$ is designed to be a compact, interpretable control surface at conditioning time.

By itself, this regression objective tends to cluster cells with similar measurements close together in the latent space, which can collapse fine distinctions between cells. To prevent this, we add a dispersion (entropy-maximization) term. Given a batch of latents $\{z_{i}\}_{i=1}^{B}$ of size $B$, we define $\mathcal{L}_\mathrm{Entropy} = - \frac{1}{B(B-1)}\sum_{i=1}^{B}\sum_{j\neq i} ||z_i-z_j||^2$, the negative mean pairwise squared distance. This acts as a tractable proxy for latent entropy and encourages latents to spread out. While optimizing a negative squared distance alone could lead to divergence, the standard VAE KL term ($\mathcal{L}_\mathrm{KL}$) acts as a bounding prior that pulls latents toward a standard Gaussian. This dynamic equilibrium prevents global fragmentation while mitigating local mode collapse, ensuring that biologically distinct cells maintain appropriate distances without drifting to infinity. $\mathcal{L}_\mathrm{Entropy}$ is therefore a soft anti-collapse regularizer rather than a hard separation constraint: it competes against $\mathcal{L}_\mathrm{Bio}$, which actively pulls cells with similar measurements together. The equilibrium between these opposing forces preserves measurement-based similarity locally while preventing global collapse. The final BioVAE objective is:
\begin{equation}
    \mathcal{L}_\mathrm{BioVAE} = \mathcal{L}_\mathrm{VAE} + \lambda_\mathrm{Bio}\mathcal{L}_\mathrm{Bio}+\lambda_\mathrm{Entropy}\mathcal{L}_\mathrm{Entropy}.
\end{equation}

\subsection{Biological Conditioning for Latent Diffusion Models}
We now plug the BioVAE encoder into a Latent Diffusion Model (LDM)~\cite{ldm} to generate new single-cell images. The pre-trained BioVAE acts as a fixed encoder, projecting each image into a compact latent $z$. A denoising network $\epsilon_\theta$ then learns to iteratively remove Gaussian noise $\epsilon$ from the latent, and we condition the denoising network on the biological signal that the user provides. We compare two choices of conditioning signal: the full 61-dimensional measurement vector $v$, and the 15 aggregated semantic axes $s$. The first uses every measurement directly; the second uses only the aggregated, intuitive controls.

The denoising network is trained with the standard conditional latent diffusion objective: $\mathcal{L}_\mathrm{LDM} = \mathbb{E}_{z, t, \epsilon \sim \mathcal{N}(0,1)} \left[ ||\epsilon - \epsilon_\theta(z_t, t, s)||_2^2 \right]$, where $z_t$ is the noisy latent at timestep $t$, and $s$ (or alternatively $v$) is the conditioning signal. Because the BioVAE latent and the semantic axes are both aligned with the same set of measurements, moving along any axis at conditioning time produces a consistent and interpretable change in the generated cell.

\section{Experiments}

We evaluated our proposed single-cell generation pipeline on the LIVECell dataset~\cite{livecell} through three main experiments. First, we ablate the loss components of BioVAE to assess their impact on image reconstruction quality and on how well the latent space is organized along biological measurements. Second, we evaluate the generative quality of the single-cell LDM, comparing conditioning on the raw 61-dimensional measurements against conditioning on the 15 semantic axes, and complement this with a blind study in which biologists assess the realism of generated cells. Finally, we evaluate per-axis controllability through traversal experiments, in which we vary a single semantic axis at a time and measure whether the generated cells change in the expected direction.

\subsection{Experimental Setup}
\noindent\textbf{Dataset:} We isolated individual cells using their respective annotation masks, setting backgrounds to black and removing edge fragments. At the instance level, the LIVECell dataset yields $64 \times 64$ grayscale crops partitioned into 928k training, 166k validation, and 411k test images.

\noindent\textbf{Evaluation Metrics:} To evaluate \textit{reconstruction fidelity}, we used MSE, PSNR, SSIM, and rFID. MSE and PSNR assessed pixel-level accuracy, SSIM evaluated structural consistency, and rFID (reconstruction FID~\cite{FID}) captured perceptual realism using deep features. To assess \textit{latent space quality}, we measured $k$-NN neighborhood preservation and linear SVM classification accuracy for cell types. For the \textit{generative performance} of our diffusion model, we implemented domain-specialized FID~\cite{FID}, MMD~\cite{mmd}, Precision, and Recall~\cite{precision_recall} for a quantitative measurement.

\noindent\textbf{BioVAE:} Our foundational encoder was based on the Stable Diffusion VAE~\cite{ldm,vq-vae,vq-vae2} architecture with a downscale factor of 8 and a latent shape of $8 \times 8 \times 4$. Training used the Adam optimizer with learning rate of $1.0 \times 10^{-4}$ for 500k iterations with a batch size of 64. Standard loss weights were set to $\lambda_\mathrm{Recon}=1$, $\lambda_\mathrm{KL}=10^{-6}$, and $\lambda_\mathrm{GAN}=1$. BioVAE additionally applied $\lambda_\mathrm{Bio}=1$ and $\lambda_\mathrm{Entropy}=1$.

\noindent\textbf{Single-Cell Diffusion:} We trained a conditional LDM for 500K iterations using a batch size of 64 and the Adam optimizer with a learning rate of $1\times10^{-4}$. Biological conditioning (measurements $v$ or semantic axes $s$) was integrated via FiLM~\cite{film}. At training, we used 1000 steps for the linear noise scheduler, and we synthesized synthetic single-cells using a DDPM~\cite{ddpm} sampler.

\subsection{Implementation Details}
\noindent\textbf{Boundary handling.} Cells were cropped using their segmentation masks with the background set to zero, then padded to $64\times64$. To prevent the model from latching onto sharp mask boundaries as a shortcut, we applied a 2-pixel Gaussian feathering along the mask edge during training. We did not observe artifacts at the boundary in generated samples (Fig.~\ref{fig:cells_gen}).

\noindent\textbf{Domain-specialized image quality evaluation.} We used a DINOv2 ViT-S/14 backbone fine-tuned on LIVECell single-cell crops for 50 epochs with a self-supervised iBOT objective, following the default setup adapted to low-resolution images~\cite{dinov2}. The quantitative metrics were computed in the resulting 384-d embedding space using 10k real and 10k generated cells per cell line, stratified.

\subsection{Evaluation of BioVAE}
Starting from the VAE used in latent diffusion models (LDM), we performed an ablation of the biological and entropy regularization terms. We evaluated reconstruction fidelity (MSE, SSIM, PSNR, rFID), morphodynamic feature reconstruction MSE (Morph. MSE), and latent topology via $k$-NN neighborhood preservation \cite{kNN_preservation} and linear SVM classification on encoder latents (Tables~\ref{tab:quant_eval}–~\ref{tab:latent_structure}).

\begin{table}[t]
    \centering
    \caption{Quantitative ablation of BioVAE loss components. Metrics include MSE ($\times10^{-4}$). \textbf{Bold} and \underline{underlined} values indicate the best and second-best results, respectively. BioVAE attains the best pixel-level reconstruction (MSE, SSIM, PSNR) while remaining comparable on rFID; the entropy term trades a small increase in descriptor (Morph.) MSE for improved reconstruction and latent structure (Table~\ref{tab:latent_structure}).}
    \vspace{1.5mm}
    \begin{tabular}{l|ccccc}
        \toprule
        Method & MSE$\downarrow\ $ & SSIM$\uparrow$ & PSNR$\uparrow$ & rFID$\downarrow$ & Morph. MSE$\downarrow$  \\ \hline
        VAE~\cite{ldm} & 0.805 & \underline{0.993} & 46.6 & {\bf 0.151} & 6.748 \\
        \textbf{BioVAE} & {\bf 0.600} & {\bf 0.994} & {\bf 49.8} & \underline{0.161} & 6.816 \\
        BioVAE w/o $\mathcal{L}_\mathrm{Bio}$ & 0.795 & 0.987 & 46.7 & 0.191 & \underline{6.699} \\
        BioVAE w/o $\mathcal{L}_\mathrm{Entropy}$ & \underline{0.630} & \underline{0.993} & \underline{49.3} & 0.341 & {\bf 6.617} \\
        \bottomrule
    \end{tabular}
    \label{tab:quant_eval}
\end{table}

\begin{table}[t]
\centering
\caption{BioVAE latent structure. kNN neighborhood preservation against the morphodynamic space and cell type classification accuracy.}
\vspace{1.5mm}
\label{tab:latent_structure}
\begin{tabular}{l|ccc|c}
\toprule
Model & K=5 & K=10 & K=20 & SVM Acc \\
\hline
Chance & 0.00003 & 0.00006 & 0.00012 & 17.8\% \\
VAE & 0.0056 & 0.0040 & 0.0029 & 70.5\% \\
\textbf{BioVAE} & \textbf{0.0058} & \textbf{0.0041} & \textbf{0.0031} & \textbf{71.2\%} \\
Rel.\ gain & +3.6\% & +2.5\% & +6.9\% & +1\% ($p<0.001$) \\
\bottomrule
\end{tabular}
\end{table}

BioVAE improves reconstruction fidelity over the baseline VAE (MSE 0.60 vs 0.81; PSNR 49.8 vs 46.6) while maintaining comparable rFID, indicating that biological regularization does not degrade image quality. Adding the entropy term slightly increases descriptor (Morph.) MSE relative to the variant without it (6.82 vs 6.62), reflecting the designed competition between the descriptor-regression force, which pulls biologically similar latents together, and the dispersion force, which spreads them apart. In exchange, the full BioVAE achieves a better balance between reconstruction accuracy and latent structure, providing a more suitable representation for controllable generation. We further analyze the latent topology using k-NN neighborhood preservation and linear SVM classification accuracy. Neighborhood preservation increases consistently across all $k$, in which it measures the overlap between nearest neighbors in the latent space and those in the morphodynamic descriptor space. Although absolute values remain small due to the large evaluation set ($>$160k cells), relative gains indicate improved phenotype-aware organization. Linear classification accuracy also increases by $\sim$1\% ($p<0.001$). Entropy regularization contributes to this effect by discouraging collapsed embeddings, while the Bio loss promotes a more structured latent geometry, supporting its use as the encoder backbone for conditional diffusion.

\subsection{Generated Quality} \label{gen_qual}
We evaluated four single-cell LDMs combining BioVAE or baseline VAE encoders with the full high-dimensional morphodynamic descriptor vector $v$ or semantic axes $s$, measuring image quality metrics (FID \cite{FID}, MMD \cite{mmd}, Precision, and Recall \cite{precision_recall}) using a DINOv2 model fine-tuned on single-cell data \cite{dinov2}, as well as Morph. MSE on the LIVECell validation split (Table~\ref{tab:sd_results}).

\begin{table}[t]
\centering
\caption{Quantitative assessment of single-cell diffusion under different conditional regimes.}
\label{tab:sd_results}
\begin{tabular}{l|ccccc}
\toprule
Regimen                           & FID$\downarrow$ & MMD$\downarrow$ & Prec.$\uparrow$ & Recall$\uparrow$ & Morph. MSE$\downarrow$ \\ \midrule
VAE + Morph.                      & 2.04            & 0.0047          & 0.872               & 0.843            & \textbf{0.864}  \\
BioVAE + Morph.                   & 2.12            & 0.0047          & \textbf{0.887}      & 0.834            & {\ul 0.876}     \\
VAE + Sem. Ax.                    & {\ul 0.974}     & {\ul 0.0015}    & 0.875               & \textbf{0.86}    & 0.944           \\
\textbf{BioVAE + Sem. Ax. (Ours)} & \textbf{0.933}  & \textbf{0.0014} & {\ul 0.880}         & {\ul 0.854}      & 0.925 \\ \bottomrule
\end{tabular}
\end{table}

\begin{figure}[t]
    \centering
    \includegraphics[angle=90, width=\linewidth]{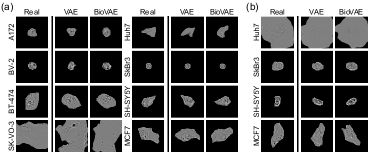}
    \caption{Generated images from VAE and BioVAE LDM with semantic axes. (a) BioVAE LDM preserves morphological features more faithfully than VAE LDM. (b) Challenging examples with low organelle preservation.}
    \label{fig:cells_gen}
\end{figure}

Semantic axes consistently outperform morphodynamic descriptors, approximately halving FID (0.93--0.97 vs 2.04--2.12), with BioVAE + semantic axes achieving the best performance (FID 0.93, MSE 0.925). Interestingly, while BioVAE underperforms the baseline VAE when conditioned on the highly correlated 61-dimensional vector, it demonstrates superior fidelity when paired with the 15-dimensional semantic axes. This suggests that because BioVAE's latent space already strongly encodes phenotypic structure, conditioning directly on the redundant 61-dimensional vector causes information interference. The compact, biologically meaningful 15-dimensional semantic axes alleviate this redundancy, creating a favorable synergy with the biologically regularized latent space. Fig.~\ref{fig:cells_gen}~(a) shows representative generated cells, and Fig.~\ref{fig:cells_gen}~(b) highlights challenging cases with partial organelle detail loss.

Notably, the ``VAE + morphodynamic (61-d)'' configuration in Table~\ref{tab:sd_results} instantiates the core design choices of prior descriptor-to-image methods --- an unstructured latent space conditioned on a high-dimensional, correlated feature vector~\cite{CP2Img}. Our full model (BioVAE + semantic axes) differs precisely along the two axes we ablate: latent regularization and conditioning structure. The 2$\times$2 ablations therefore double as a controlled comparison against that design paradigm, under a single shared encoder, evaluator, and protocol.

\subsection{Blind Study with Biologists}
\begin{figure}[t]
    \centering
    \includegraphics[width=0.8\linewidth]{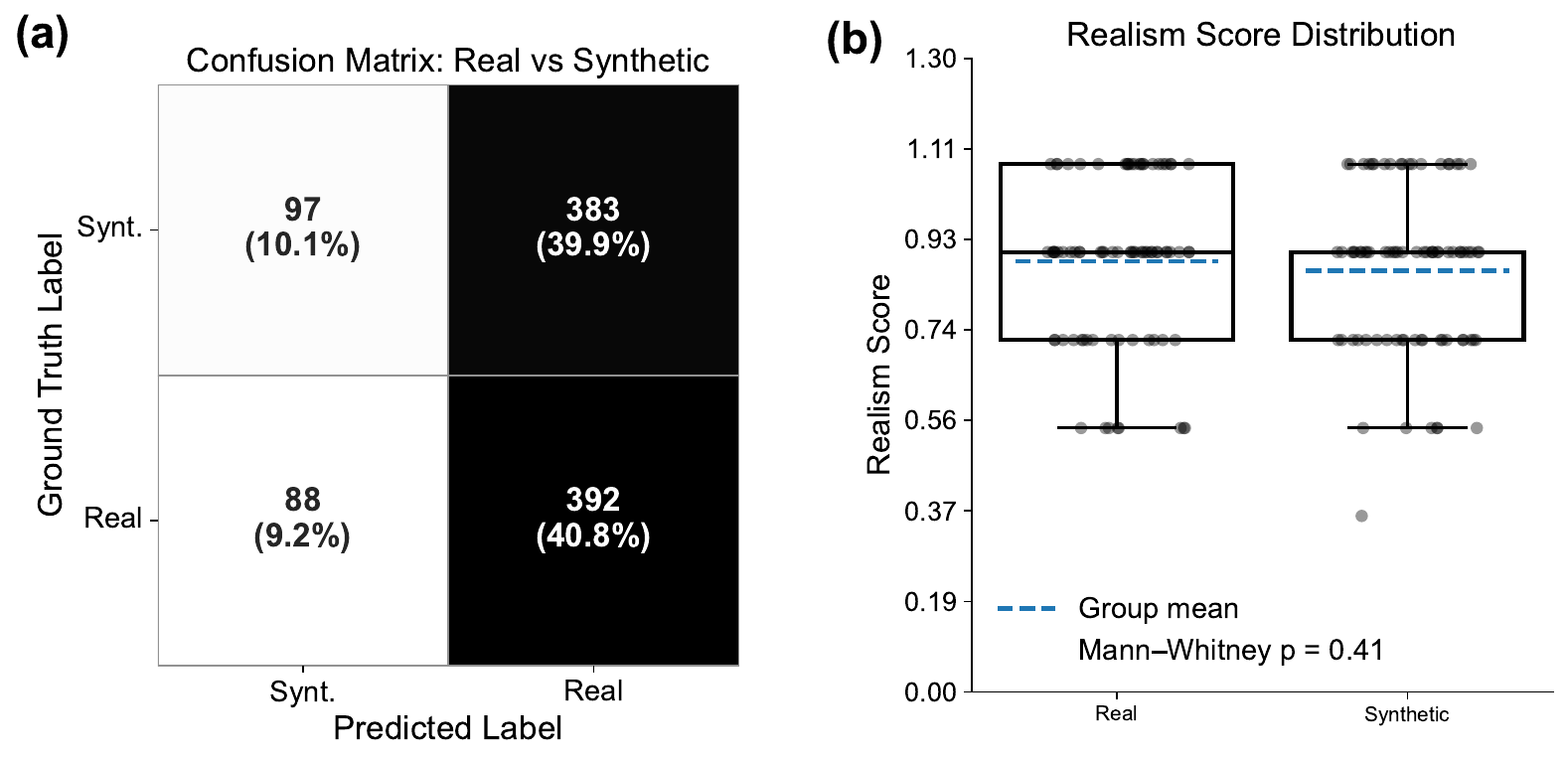}
    \caption{Biology expert evaluation. (a) Confusion matrix for real versus synthetic classification. (b) Distribution of realism scores for real and synthetic images.}
    \label{fig:blintest}
\end{figure}

A blind study with six biologists evaluated 960 cell instances across eight cell types. Participants classified each cell as real or synthetic, predicted its cell type, and rated task difficulty on a 0--4 scale. Overall discrimination accuracy was 50.9\% (AUC = 0.509), which was not significantly different from random guessing (binomial $p=0.58$). As shown in Fig.~\ref{fig:blintest}~(a), synthetic samples were frequently perceived as real: only 20.2\% of synthetic cells were correctly identified as synthetic, whereas 81.7\% of real cells were correctly classified as real. To quantify perceived realism at the sample level, we defined a realism score as the fraction of observers classifying a cell as real. Real and synthetic cells exhibited comparable realism score distributions, with no significant difference between groups (Mann--Whitney $p=0.41$; Fig.~\ref{fig:blintest}~(b)). Observers generally rated the task as difficult (mean difficulty of 3.17/4), and their ratings were not significantly correlated with discrimination accuracy ($r=0.14$, $p=0.34$). Together, these results indicate that generated single-cell images are visually realistic and preserve morphology consistent with the target phenotypes. Combined with the quantitative analyses above, the study supports the ability of semantic-axis conditioning on BioVAE latents to produce high-quality, phenotype-aligned single-cell images while maintaining biological interpretability.

\subsection{Controllability via Axis Traversal}\label{sec:control}
Beyond image quality, we explored the controllability properties of the BioVAE + semantic axes latent diffusion model. Unlike the biological feature representation, the semantic axes provide a compact and interpretable phenotype space that enables controllability analysis through biologically meaningful semantic traversals. For this experiment, we sampled 100 representative validation cells, referred to as anchors, and generated semantic traversals by modifying one semantic axis at a time while keeping the remaining axes fixed. The semantic axes were designed to prioritize biological interpretability rather than orthogonal latent decomposition; therefore, the controllability analysis should be interpreted as biologically consistent semantic modulation within a correlated phenotype manifold rather than fully disentangled control.

To quantitatively evaluate semantic controllability, we measured three complementary properties: target controllability, control gain, and selectivity. Target controllability was estimated using the Spearman correlation between requested and measured semantic values. Control gain was estimated using the slope of a linear regression between requested and generated semantic values. Finally, selectivity measures the relative dominance of the target semantic response compared to the average response of the non-target semantic axes.

\begin{figure}[t]
    \centering
    \includegraphics[width=0.9\linewidth]{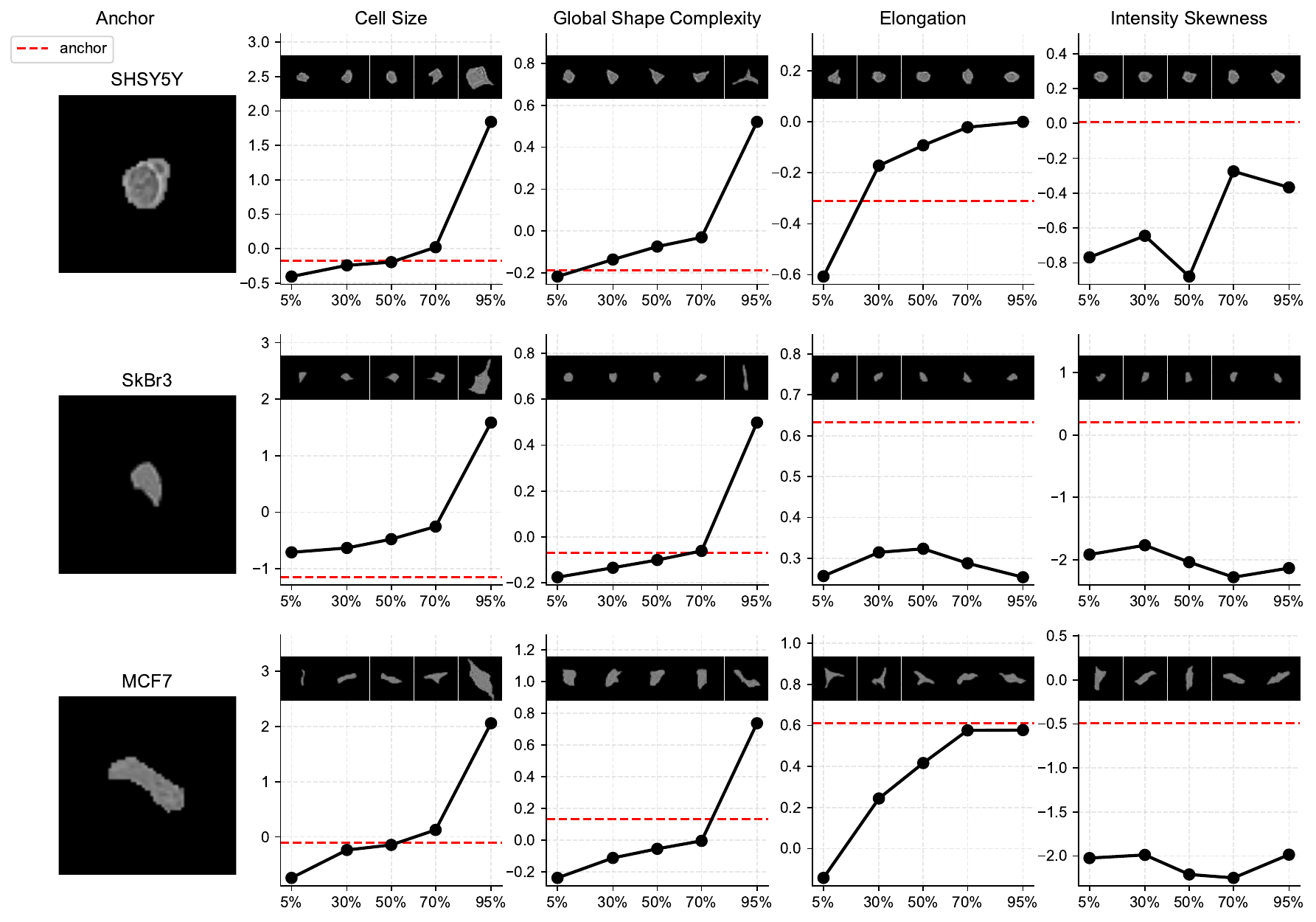}
    \caption{Representative semantic traversals generated from three validation anchor cells. Each line plot shows the measured semantic-axis response across the traversal range, with their corresponding generated cell images. Dashed red lines indicate the semantic value of the original anchor cell.}
    \label{fig:traversal_gen}
\end{figure}

Fig.~\ref{fig:traversal_gen} presents representative semantic traversals from a shared anchor cell. The quantitative results are summarized in the Supplementary Sec.~C, with additional anchor examples.

The traversal results reveal multiple controllability regimes across the semantic space. Cell size, global shape complexity, and symmetry exhibited high target controllability, strong control gain, and moderate selectivity, suggesting that the model can effectively manipulate global morphological structure. Elongation, boundary irregularity, and intensity contrast also presented high target controllability but lower control gain, indicating compressed semantic responses relative to the requested traversal range. We attribute this behavior to the model remaining close to the learned biological manifold and preserving biologically plausible phenotype combinations during generation, consistent with the high precision and recall observed in Section \ref{gen_qual}.

In contrast, shape compactness and granularity exhibited weak target controllability, low control gain, and low selectivity, suggesting limited semantic responsiveness. Interestingly, axes such as edge strength, texture heterogeneity, and intensity skewness presented low target controllability but comparatively high selectivity. This behavior suggests that the model can induce localized semantic perturbations for these descriptors, although without consistently following the requested traversal direction. These axes are strongly correlated with other semantic descriptors and partially reuse related biological features, making isolated manipulation intrinsically more difficult.

\section{Discussion and Limitations}

\noindent\textbf{The unit of generation and its scope.}
Whole-image and single-cell generation address complementary aspects of microscopy. Whole-image models capture spatial organization and cell-to-cell context, whereas single-cell models provide direct control over the appearance of individual cells. Our framework focuses on the latter and therefore does not model interactions between neighboring cells or larger tissue context. In addition, our evaluation is restricted to LIVECell, which contains phase-contrast images from eight cell lines. Extending the framework to fluorescence, brightfield, 3D imaging, and multi-cellular scenes is an important direction for future work.

\noindent\textbf{Why the input and latent representation should be designed together.}
A central finding is that biologically structuring the latent representation alone is insufficient. BioVAE does not improve generation when paired with the full 61-dimensional measurement vector; its benefit emerges when conditioning is also organized through the 15 semantic axes (Sec.~4.4). This suggests that controllability is a property of the overall pipeline rather than of a single component. More broadly, our results indicate that simple biological priors can improve not only interpretability but also representation structure and conditional generation quality (Tables~2--3).

\noindent\textbf{Scope of the semantic axes.}
The semantic axes provide a compact set of named, measurable, and traversable controls over cell appearance, but they are not a complete description of cellular phenotype. Fine-grained organelle structure, sub-cellular spatial patterns, and biological state transitions are not explicitly represented. Moreover, the axes are manually constructed from standard measurements, so their coverage depends on what those measurements capture. Data-driven extensions using richer assays such as Cell Painting may provide additional biologically meaningful control directions.

\noindent\textbf{External comparison.}
Our evaluation relies primarily on controlled internal ablations rather than direct comparison with external single-cell generators, because existing methods do not provide compatible single-cell-level metrics, feature extractors, or generation protocols on LIVECell. Establishing a common benchmark for controllable single-cell generation therefore remains an important community-level direction.

\section{Conclusion}
We presented a framework for microscopy generation that operates at the level of individual cells and designs the user-facing controls and the model's internal representation together, so that both are organized along the same set of biologically meaningful directions. On the input side, we aggregate standard cell measurements into a small set of semantic axes that are each named, measurable, and traversable. On the model side, BioVAE aligns its latent space with the same measurements. On LIVECell, this co-design substantially improves generation quality over conditioning on raw high-dimensional measurements, produces single cells that biologists cannot reliably distinguish from real ones in a blind study, and supports stable per-axis control verified through traversal experiments. We see controllable single-cell generation as a complementary building block alongside whole-image approaches, one that connects classical quantitative cell biology to modern generative modeling and opens a research direction toward biologically grounded, controllable microscopy synthesis.

\section*{Acknowledgment}
This study was partially funded by Sartorius AI Lab (SAIL), a collaboration between Sartorius (\url{https://www.sartorius.com}) and the German Research Center for Artificial Intelligence (DFKI) (\url{https://www.dfki.de}). The funders had no role in study design, data collection and analysis, decision to publish, or preparation of the manuscript.

This work was also supported by Japan Science and Technology Agency (JST) as part of Adopting Sustainable Partnerships for Innovative Research Ecosystem (ASPIRE), Grant Number JPMJAP2403, and Japan Society for the Promotion of Science (JSPS) KAKENHI Grant Numbers JP24KJ1805 and  JP25K22846.

\bibliography{egbib}

\clearpage
\appendix
\setcounter{figure}{0}
\setcounter{table}{0}
\setcounter{equation}{0}
\renewcommand{\thefigure}{S\arabic{figure}}
\renewcommand{\thetable}{S\arabic{table}}
\renewcommand{\theequation}{S\arabic{equation}}
\renewcommand{\theHfigure}{supp.\arabic{figure}}
\renewcommand{\theHtable}{supp.\arabic{table}}
\renewcommand{\theHequation}{supp.\arabic{equation}}
\section*{Supplementary Material}

\section{Morphodynamic descriptors}
\label{sec:morpho_desc}
The morphodynamic descriptors $v \in \mathbb{R}^{61}$ are categorized and defined as:
\subsection{Morphology}

\begin{enumerate}
    \item \textbf{Area}: Number of pixels belonging to the segmented object region.

        \begin{displaymath}
          A = N_{\text{pixels}}
        \end{displaymath}

        where $N_{\text{pixels}}$ is the total number of foreground pixels.
    \item \textbf{Perimeter}: Number of pixels located on the boundary of the segmented object.

        \begin{displaymath}
          P = N_{\text{boundary}}
        \end{displaymath}

        where $N_{\text{boundary}}$ denotes the number of boundary pixels.
    \item \textbf{Equivalent diameter}: Diameter of a circle with the same number of pixels as the segmented object.

        \begin{displaymath}
          D_{eq} = \sqrt{\frac{4A}{\pi}}
        \end{displaymath}
    \item \textbf{Major and minor axis length}: Lengths of the principal axes of the ellipse fitted to the object using second-order image moments.

        \begin{displaymath}
          L_{\text{major}} = 2a \quad ; \quad L_{\text{minor}} = 2b
        \end{displaymath}

        where $a$ and $b$ are the semi-major and semi-minor ellipse axes estimated from the pixel distribution.
    \item \textbf{Aspect ratio}: Ratio between the major and minor axis lengths.

        \begin{displaymath}
          AR = \frac{L_{\text{major}}}{L_{\text{minor}}}
        \end{displaymath}
    \item \textbf{Eccentricity}: Measures the deviation of the fitted ellipse from a circular shape.

        \begin{displaymath}
          e = \sqrt{1 - \frac{b^2}{a^2}}
        \end{displaymath}

        where $a$ and $b$ are the semi-major and semi-minor ellipse radii.
    \item \textbf{Solidity}: Fraction of convex hull pixels occupied by the object.

        \begin{displaymath}
          S = \frac{A}{A_{\text{convex}}}
        \end{displaymath}

        where:
        \begin{itemize}
            \item $A$ is the object area in pixels,
            \item $A_{\text{convex}}$ is the convex hull area in pixels.
        \end{itemize}
    \item \textbf{Circularity}: Measures how closely the object resembles a perfect circle.

        \begin{displaymath}
          C = \frac{4\pi A}{P^2}
        \end{displaymath}

        where:
        \begin{itemize}
            \item $A$ is the object area,
            \item $P$ is the estimated perimeter.
        \end{itemize}
    \item \textbf{Convex Area}: Number of pixels contained within the convex hull of the object.

        \begin{displaymath}
          A_{\text{convex}} = N_{\text{convex pixels}}
        \end{displaymath}
    \item \textbf{Extent}: Fraction of pixels within the bounding box occupied by the object.
        \begin{displaymath}
          E = \frac{A}{A_{\text{bbox}}}
        \end{displaymath}

        where:
        \begin{itemize}
            \item $A$ is the object area,
            \item $A_{\text{bbox}}$ is the bounding box area in pixels.
        \end{itemize}
    \item \textbf{First 9 Zernike moments}: Rotation-invariant orthogonal shape descriptors computed over the binary object mask mapped to the unit disk.

        \begin{displaymath}
          Z_{nm} = \frac{n+1}{\pi}
            \sum_x \sum_y I(x,y),V_{nm}^{*}(x,y)
        \end{displaymath}

        where:
        \begin{itemize}
            \item $I(x,y)$ is the binary pixel intensity,
            \item $V_{nm}(x,y)$ is the Zernike polynomial,
            \item $(\cdot)^*$ denotes the complex conjugate.
        \end{itemize}

        The first 9 moments correspond to the lowest-order polynomial combinations and capture coarse global shape information.
\end{enumerate}

\subsection{Pixel Intensity}
\begin{enumerate}
    \item \textbf{Mean intensity}: Average pixel intensity within the segmented object mask.

        \begin{displaymath}
          \mu = \frac{1}{N}\sum_{i=1}^{N} I_i
        \end{displaymath}
    \item \textbf{Standard deviation of intensity}: Measures the variability of pixel intensities within the object region.

        \begin{displaymath}
          \sigma = \sqrt{\frac{1}{N}\sum_{i=1}^{N}(I_i - \mu)^2}
        \end{displaymath}

        where $\mu$ is the mean intensity.
    \item \textbf{Minimum intensity}: Lowest pixel intensity value inside the segmented region.

        \begin{displaymath}
          I_{\min} = \min_{i=1,\dots,N} I_i
        \end{displaymath}

    \item \textbf{Maximum intensity}:

        \begin{displaymath}
          I_{\max} = \max_{i=1,\dots,N} I_i
        \end{displaymath}

    \item \textbf{Intensity sum}: Total accumulated intensity inside the object mask.

        \begin{displaymath}
          I_{\text{sum}} = \sum_{i=1}^{N} I_i
        \end{displaymath}

    \item \textbf{Median intensity}: Median value of the intensity distribution within the segmented object.

        \begin{displaymath}
          I_{\text{median}} = \operatorname{median}(I_1, I_2, \dots, I_N)
        \end{displaymath}

    \item \textbf{n-th percentile intensity}: Intensity value below which $n\%$ of foreground pixel intensities are found.

        \begin{displaymath}
          I_{n} = P_{n}(I)
        \end{displaymath}

        where $P_{n}$ denotes the n-th percentile operator with $n \in \{5, 25, 75, 95\}$.
\end{enumerate}

\subsection{Texture}
\begin{enumerate}
    \item \textbf{Haralick GLCM features (13)}: Texture descriptors computed from the Gray-Level Co-occurrence Matrix (GLCM), which captures the frequency of neighboring gray-level intensity pairs within the segmented object region.

    The GLCM is defined as:
        \begin{displaymath}
          P(i,j)
        \end{displaymath}

    where $P(i,j)$ represents the normalized probability of observing gray levels $i$ and $j$ at a predefined spatial offset.

    From the GLCM, 13 Haralick texture descriptors were extracted, including:

    \begin{itemize}
        \item $H_1$: Angular Second Moment (Energy)
        \item $H_2$: Contrast
        \item $H_3$: Correlation
        \item $H_4$: Variance
        \item $H_5$: Inverse Difference Moment (Homogeneity)
        \item $H_6$: Sum Average
        \item $H_7$: Sum Variance
        \item $H_8$: Sum Entropy
        \item $H_9$: Entropy
        \item $H_{10}$: Difference Variance
        \item $H_{11}$: Difference Entropy
        \item $H_{12}, H_{13}$: Information Measures of Correlation 1 and 2
    \end{itemize}

    \item \textbf{Local Binary Pattern (LBP) histogram (10 bins)}: Rotation-invariant local texture descriptor based on thresholding neighboring pixels relative to the center pixel intensity.

    The LBP code for a pixel is computed as:
        \begin{displaymath}
          \text{LBP}(x_c,y_c)=\sum_{p=0}^{P-1}s(I_p - I_c),2^p
        \end{displaymath}

    where:
        \begin{displaymath}
          s(x)=
              \begin{cases}
                  1 & x \geq 0 \\
                  0 & x < 0
              \end{cases}
        \end{displaymath}

    and:

        \begin{itemize}
            \item $I_c$ is the center pixel intensity,
            \item $I_p$ are neighboring pixel intensities,
            \item $P$ is the number of neighbors.
        \end{itemize}

    A normalized histogram with 10 bins was computed from the resulting LBP codes within the object mask.

    \item \textbf{Gradient mean and standard deviation}: Statistics computed from image intensity gradients within the segmented object region, capturing local edge strength and texture variation.

    Gradient magnitude was computed as:
        \begin{displaymath}
          G(x,y)=\sqrt{\left(\frac{\partial I}{\partial x}\right)^2 + \left(\frac{\partial I}{\partial y}\right)^2}
        \end{displaymath}

    The mean and standard deviation of gradient magnitudes were then calculated:

        \begin{displaymath}
            \mu_G = \frac{1}{N}\sum_{i=1}^{N} G_i \quad ; \quad  \sigma_G =\sqrt{\frac{1}{N}\sum_{i=1}^{N}(G_i-\mu_G)^2}
        \end{displaymath}

    where $G_i$ denotes the gradient magnitude at foreground pixel $i$.
\end{enumerate}

\subsection{Spatial distribution}

\begin{enumerate}
    \item \textbf{6-bin radial intensity profile}: Descriptor capturing the spatial distribution of pixel intensities from the object center toward its boundary. The segmented object region is divided into six concentric radial bins centered at the object centroid, and the mean intensity is computed within each bin.

    The radial distance of a foreground pixel is defined as:

        \begin{displaymath}
            r_i =\sqrt{(x_i - x_c)^2 +(y_i - y_c)^2}
        \end{displaymath}

    where:
        \begin{itemize}
            \item $(x_i,y_i)$ are the coordinates of pixel (i),
            \item $(x_c,y_c)$ is the object centroid.
        \end{itemize}

    Let $R_{\max}$ denote the maximum radial distance within the object mask. The radial interval is uniformly divided into six bins:

        \begin{displaymath}
            B_k =
              \left[
              \frac{(k-1)R_{\max}}{6},
              \frac{kR_{\max}}{6}
              \right)
        \end{displaymath}

    for $k=1,\dots,6$.

    The intensity value for each radial bin is then computed as:

        \begin{displaymath}
            r_i =\sqrt{(x_i - x_c)^2 +(y_i - y_c)^2}
        \end{displaymath}
    where:
        \begin{itemize}
            \item $I_i$ is the intensity of foreground pixel $i$,
            \item $N_k$ is the number of pixels within radial bin $B_k$.
        \end{itemize}
\end{enumerate}

\section{Semantic axes}
    The semantic axes $s \in \mathbb{R}^{15}$ is the aggregation of the morphodynamic descriptors into biologically interpretable phenotype properties. To avoid bias towards big magnitudes, we normalized the features with the z-score:

        \begin{displaymath}
            z(x)=\frac{x-\mu_x}{\sigma_x}
        \end{displaymath}

\subsection{Morphology}
\begin{enumerate}
    \item \textbf{Cell size}: How large is the object overall.
        \begin{displaymath}
            Cell Size=\frac{z(A),z(D_{eq}),z(A_{convex}),z(L_{major}),z(L_{minor})}{5}
        \end{displaymath}
    \item \textbf{Elongation}: How stretched or directional is the object.
        \begin{displaymath}
            Elongation=\frac{z(AR),z(e)}{2}
        \end{displaymath}
    \item \textbf{Shape compactness}: How close is the shape to a circular and compact object.
        \begin{displaymath}
            Compactness=\frac{z(C),z(S)}{2}
        \end{displaymath}
    \item \textbf{Boundary irregularity}: How smooth or jagged is the cell outline
        \begin{displaymath}
           BoundaryIrregularity = \frac{z\left(\frac{P}{\sqrt{A}}\right), -z(S), -z(C)}{3}
        \end{displaymath}
    \item \textbf{Global shape complexity}: How complex is the global object shape beyond simple geometry.
        \begin{displaymath}
           GlobalShapeComplexity = \frac{1}{9} \sum_{\substack{(n,m) \ n>0}} z(|Z_{nm}|)
        \end{displaymath}
    \item \textbf{Symmetry}: Captures global rotational balance and shape regularity.

        \begin{displaymath}
           \mathrm{Symmetry}=\frac{z(|Z_2|)+z(|Z_4|)+z(|Z_6|)+z(|Z_8|)}{4}
        \end{displaymath}
\end{enumerate}

\subsection{Pixel Intensity}
\begin{enumerate}
    \item \textbf{Overall signal level}: How bright is the object
        \begin{displaymath}
           \mathrm{OverallSignal}=\frac{z(|\mu|)+z(|I_{median}|)}{2}
        \end{displaymath}
    \item \textbf{Intensity contrast}: How much variation exists within the cell
        \begin{displaymath}
           \mathrm{Contrast}=z(\sigma)
        \end{displaymath}
    \item \textbf{Intensity skewness}: Is the signal dominated by hotspots or evenly distributed?
        \begin{displaymath}
           \mathrm{Skewness}=z\left(\frac{I_{95}-2I_{50}+I_5}{I_{95}-I_5}\right)
        \end{displaymath}
\end{enumerate}

\subsection{Texture}
\begin{enumerate}
    \item \textbf{Texture heterogeneity}: How heterogeneous is the intracellular texture.
        \begin{displaymath}
           \mathrm{TextureHeterogeneity}=\frac{z(H_9)+z(H_2)+z(\sigma_G)}{3}
        \end{displaymath}

    \item \textbf{Texture smoothness}: How smooth or uniform the texture is

        \begin{displaymath}
           \mathrm{TextureSmoothness}=z(H_5)
        \end{displaymath}
    \item \textbf{Granularity}: Finds fine-scale repeating patterns
        \begin{displaymath}
           H_{\mathrm{LBP}}=z\left(-\sum_{i=1}^{10}p_i \log p_i\right)
        \end{displaymath}
    \item \textbf{Edge strength}: How strong are the internal boundaries and edges
        \begin{displaymath}
           \mathrm{EdgeStrength}=\frac{z(\mu_G)+z(\sigma_G)}{2}
        \end{displaymath}
\end{enumerate}

\subsection{Spatial Distribution}
\begin{enumerate}
    \item \textbf{Central Enrichment}: Tells if the signal is concentrated toward the center or edges
        \begin{displaymath}
           \mathrm{CentralEnrichment}=z\left(\left(\frac{r_0+r_1+r_2}{3}\right) -\left(\frac{r_3+r_4+r_5}{3}\right)\right)
        \end{displaymath}

    \item \textbf{Radial uniformity}: How evenly distributed is the signal radially
        \begin{displaymath}
           \sigma_r=\sqrt{\frac{1}{6}\sum_{k=1}^{6}\left(r_k - \frac{1}{6}\sum_{k=1}^{6} r_k)\right)^2}
        \end{displaymath}
\end{enumerate}

\newpage

\subsection{Spearman Correlation}
\begin{figure}[h]
    \centering
    \includegraphics[width=\linewidth]{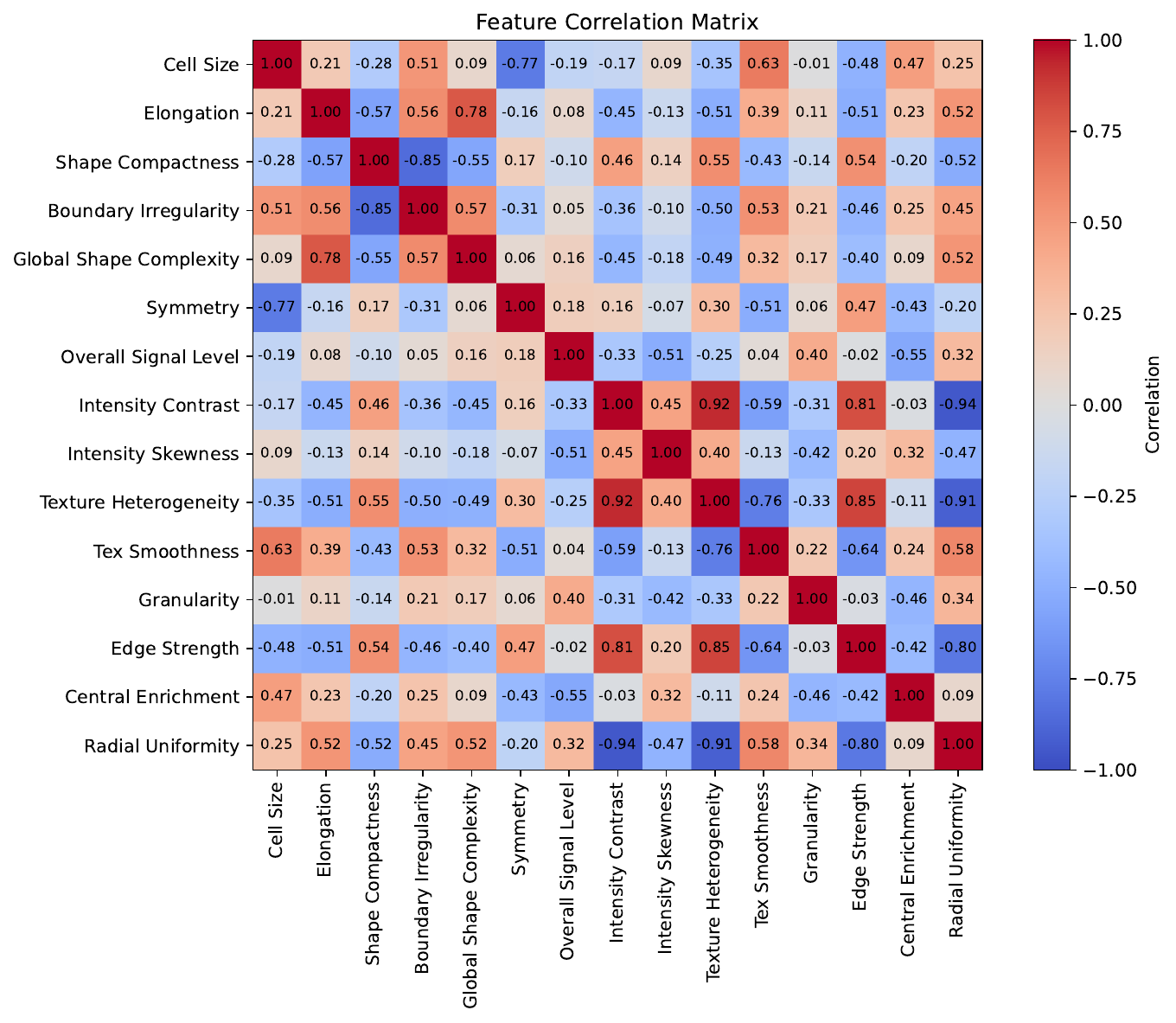}
    \caption{Spearman correlation matrix of the semantic axes in the train data split. Several semantic axes exhibit structured biological dependencies due to shared underlying descriptors and correlated phenotype properties.}
    \label{fig:supp_correlation}
\end{figure}

\newpage

\section{Semantic axes traversal}
\begin{table}[h]
    \centering
    \caption{Quantitative controllability analysis of the 15 semantic axes. Target controllability is measured using the Spearman correlation between requested and generated semantic values. Control gain is estimated as the slope of a linear regression between requested and generated semantic values. Selectivity is defined as the ratio between the target-axis semantic displacement and the average displacement of the non-target semantic axes, quantifying the relative dominance of the requested semantic response.
}
    \vspace{1.5mm}
    \begin{tabular}{l|ccc}
        \toprule
        Axis                      & Controllability $\uparrow$ & Gain $\uparrow$ & Selectivity $\uparrow$ \\ \midrule
        cell\_size                & 0.978                      & 0.712           & 1.544                  \\
        elongation                & 0.754                      & 0.287           & 0.957                  \\
        shape\_compactness        & 0.006                      & 0.027           & 0.448                  \\
        boundary\_irregularity    & 0.599                      & 0.199           & 1.043                  \\
        global\_shape\_complexity & 0.978                      & 0.901           & 0.945                  \\
        symmetry                  & 0.915                      & 0.643           & 1.169                  \\
        overall\_signal\_level    & 0.608                      & 0.682           & 0.339                  \\
        intensity\_contrast       & 0.605                      & 0.318           & 1.42                   \\
        intensity\_skewness       & 0.187                      & 0.213           & 1.757                  \\
        tex\_heterogeneity        & 0.39                       & 0.111           & 1.899                  \\
        tex\_smoothness           & 0.673                      & 0.305           & 1.203                  \\
        granularity               & 0.177                      & 0.064           & 0.382                  \\
        edge\_strength            & -0.211                     & -0.027          & 7.275                  \\
        central\_enrichment       & 0.732                      & 0.491           & 1.05                   \\
        radial\_uniformity        & 0.222                      & 0.138           & 0.776                  \\ \bottomrule
    \end{tabular}
\end{table}

\begin{figure}[t]
    \centering
    \includegraphics[width=\linewidth]{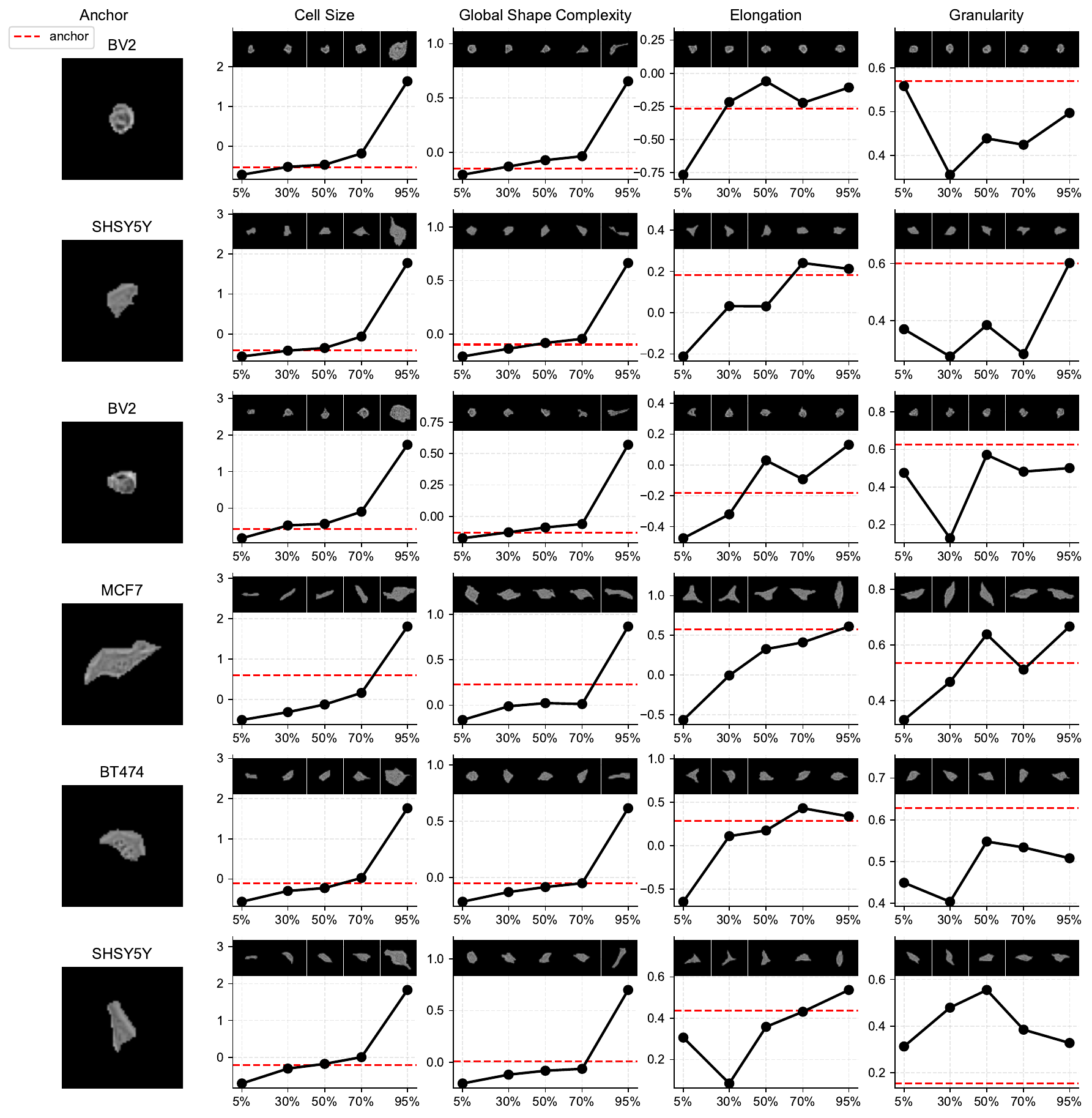}
    \caption{Representative semantic traversals generated from six validation anchor cells. Each line plot shows the measured semantic-axis response across the traversal range, with their corresponding generated cell images. Dashed red lines indicate the semantic value of the original anchor cell.}
    \label{fig:supp_traversal_gen}
\end{figure}

\section{Blind Study Details}
The blind study involved six biologists who evaluated 960 cell instances
across eight cell types. Participation was voluntary, and participants were
informed that their responses would be used for research purposes. Responses
were analyzed only in aggregate, and no personally identifiable information
was collected.

\end{document}